\documentclass[11pt,letterpaper,logo,twocolumn]{style}

\usepackage{amsmath,amssymb,amsfonts}
\usepackage{booktabs,multirow,array}
\usepackage{graphicx}
\usepackage{textcomp,url}
\usepackage{latexsym}
\usepackage{listings}
\tcbuselibrary{listings,listingsutf8,skins,breakable}
\usepackage{subcaption}
\graphicspath{{./}{figure/}}

\PublicDate{2026-06-21}
\runningtitle{HG-Bench: Multi-Page Handwritten Answer-Region Grounding}

\definecolor{headergray}{RGB}{230,230,230}
\definecolor{rowgray}{RGB}{245,245,245}
\definecolor{oursblue}{HTML}{F1E4F4}
\definecolor{gainblue}{HTML}{82318E}

\newcommand{\best}[1]{\textbf{#1}}
\newcommand{\second}[1]{\underline{#1}}
\newcommand{\gain}[1]{\textcolor{gainblue}{$\uparrow$ #1}}

\definecolor{jsonbg}{HTML}{F8F1FA}
\definecolor{jsonframe}{HTML}{CFA7D6}
\definecolor{jsontext}{HTML}{2E2A35}

\title{HG-Bench: A Benchmark for Multi-Page Handwritten Answer-Region Grounding in Automated Homework Assessment}

\author{%
  {\Authfont
    \href{https://openreview.net/profile?id=~Chuangxin_Zhao1}{\textbf{Chuangxin Zhao}} \quad
    \href{https://openreview.net/profile?id=~Boyan_Shi1}{\textbf{Boyan Shi}} \quad
    \href{https://openreview.net/profile?id=~Yanling_Wang1}{\textbf{Yanling Wang}} \quad
    \href{https://openreview.net/profile?id=~Yijian_LU1}{\textbf{Yijian LU}} \quad
    \href{https://openreview.net/profile?id=~Canran_Xiao1}{\textbf{Canran Xiao}}\par\vspace{0.35em}
    \href{https://openreview.net/profile?id=~Jiali_Chen1}{\textbf{Jiali Chen}} \quad
    \href{https://openreview.net/profile?id=~Jun_Xia1}{\textbf{Jun Xia}} \quad
    \href{https://openreview.net/profile?id=~Yan_Wang42}{\textbf{Yan Wang}} \quad
    \href{https://openreview.net/profile?id=~Ji_Qi3}{\textbf{Ji Qi}}\advisor \quad
    \href{https://openreview.net/profile?id=~Juanzi_Li1}{\textbf{Juanzi Li}}\advisor
  }\par\vspace{0.45em}
  {\Affilfont \advisor\ Corresponding authors: Ji Qi and Juanzi Li. Email: \texttt{\{qiji,lijuanzi\}@tsinghua.edu.cn}}
}


\begin{document}

\begin{abstract}
Automated homework assessment depends not only on recognizing student answers, but also on accurately locating where each answer and each intermediate reasoning step appears in noisy, multi-page handwritten work. 
This paper addresses the missing evaluation setting of \emph{page-aware, two-level answer-region grounding}: given a sequence of homework page images, a model must localize complete answer regions and their ordered step-level subregions. 
We introduce \textbf{HG-Bench}, a benchmark of $500$ human-annotated K--12 homework samples curated from a $1{,}489{,}278$-image source pool, with question-level and step-level boxes linked by a hierarchical containment constraint. 
HG-Bench is paired with a page-aware evaluation protocol that separately measures complete-answer localization ($\mathcal{F}_A$) and step-level decomposition ($\mathcal{F}_S^{\mu}$), revealing whether models truly ground the spatial structure of student reasoning rather than merely parse visible text. 
Across frontier closed-source APIs and competitive open-weight VLMs, no zero-shot system exceeds $55.22\%$ on $\mathcal{F}_A$ or $48.22\%$ on $\mathcal{F}_S^{\mu}$, while a GLM-4.6V~9B reference model fine-tuned on $\sim$10k in-domain examples reaches $74.97/72.26$. 
These results identify step-level handwritten grounding as a concrete capability gap and provide a reproducible benchmark, evaluation protocol, and trained reference point for future work on automated homework assessment. 
Project page: \url{https://hg-bench.github.io}
\end{abstract}

\maketitle

\begin{figure}[t]
  \centering
  \includegraphics[width=\linewidth]{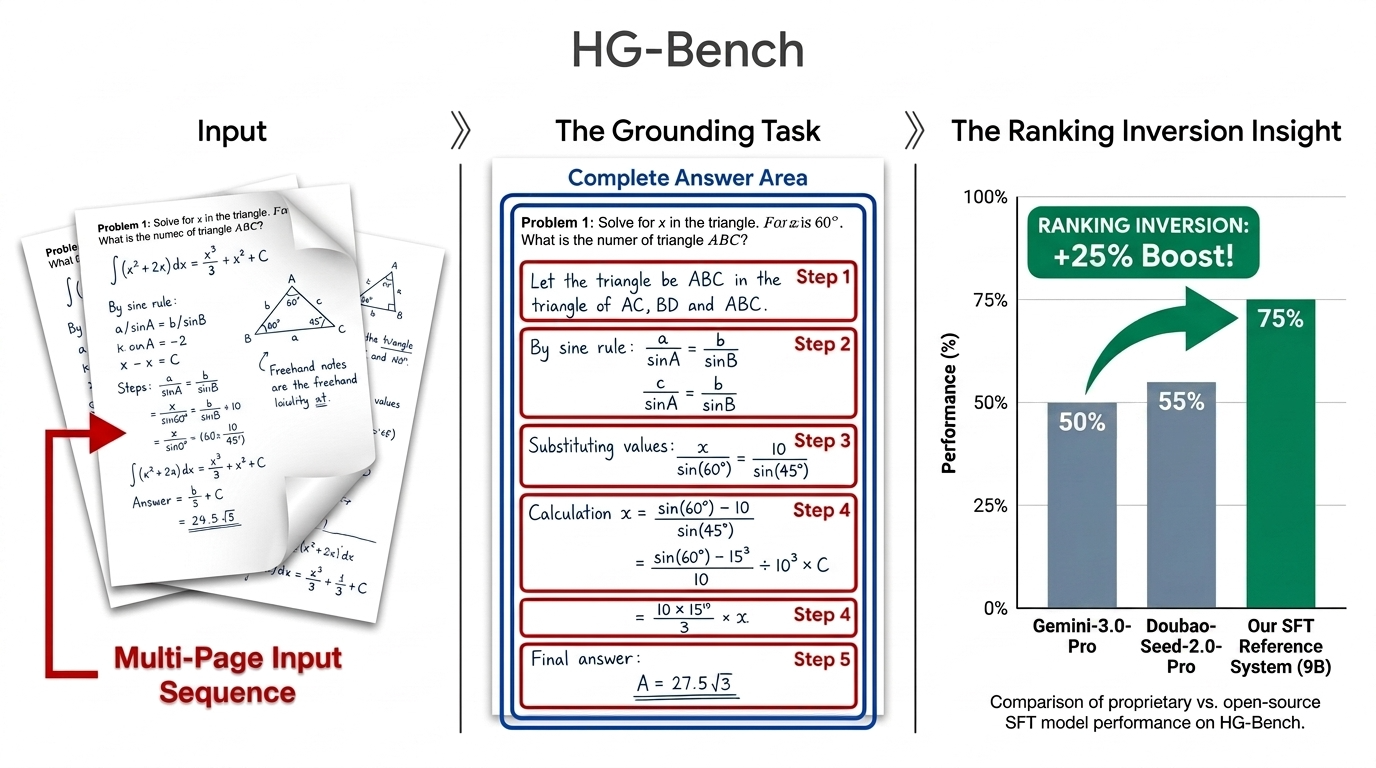}
  \caption{\textbf{HG-Bench at a glance.}
HG-Bench takes a sequence of multi-page handwritten homework images as input and requires models to produce page-aware, two-level grounding outputs: complete answer regions for each question and ordered step-level boxes for multi-step solutions. 
This setting tests whether VLMs can localize the spatial structure of student reasoning, not only recognize final answers. 
}
  \label{fig:teaser}
\end{figure}

\section{Introduction}
\label{sec:intro}

Automated homework assessment is increasingly deployed in educational settings to relieve teacher workload and to improve grading consistency. The first stage of every such pipeline is \emph{spatial
localization}: the system must identify, on each scanned page, where a student has written each answer and, for multi-step problems, where each constituent step of the derivation lies. Downstream optical
character recognition, grading, and feedback modules consume these regions, so the accuracy of the final grade is bounded above by the accuracy of upstream grounding.

\paragraph{Research gap.}
Despite rapid progress on referring-expression grounding in natural images~\cite{kazemzadeh2014refcoco,plummer2015flickr30k} and on document
understanding~\cite{mathew2021docvqa,masry2022chartqa,mathew2022infovqa}, no public benchmark measures grounding on real handwritten student work. Natural-image benchmarks evaluate a single referent per
query in editorial photographs; document-AI benchmarks recover text content but do not require spatially correct, hierarchically structured region outputs. Educational AI benchmarks such as
MathVista~\cite{lu2024mathvista} and MathVerse~\cite{zhang2024mathverse} evaluate problem solving assuming the relevant content has already been identified. The capability most relevant to real assessment
pipelines---ordered, two-level, page-aware grounding on noisy multi-page handwritten scans---remains unmeasured.

\paragraph{Core challenges.}
Real homework scans are markedly harder than the inputs of any prior grounding benchmark along four axes simultaneously: (i)~\textbf{multi-page} samples with page-aware coordinate semantics;
(ii)~\textbf{handwriting} with irregular line spacing, hand shadow, perspective skew, and student-specific stroke styles; (iii)~\textbf{two-level structure}, requiring both per-question answer regions and
ordered per-step sub-regions under hierarchical containment; and (iv)~\textbf{long-tail layout heterogeneity} across subjects (e.g., fraction derivations and matrices in mathematics versus dense paragraphs
in language subjects). A benchmark that fails to expose any one of these axes will overstate the readiness of current vision--language models (VLMs) for assessment deployment.

\paragraph{Our approach.}
We address this gap with two coordinated artefacts. First, HG-Bench: a curated, stratified, human-annotated test set of $500$ multi-page samples covering all four challenge axes, together with a page-aware
evaluation protocol that decouples question-level localization ($\mathcal{F}_A$) from step-level decomposition ($\mathcal{F}_S^{\mu}$). Second, a lightweight \emph{reference fine-tuned system} obtained by
single-stage supervised fine-tuning of an open-weight 9B VLM on $\sim$10k in-domain examples, included as a trained reference point rather than as a SOTA submission---it verifies that the benchmark is
learnable and quantifies a lower bound on dedicated-pipeline performance.

\paragraph{Key empirical findings.}
We observe a pronounced ranking inversion across paradigms (Tab.~\ref{tab:main}): the strongest closed-source frontier model attains only $55.22\%$ on $\mathcal{F}_A$ and $48.22\%$ on $\mathcal{F}_S^{\mu}$,
 whereas the reference SFT system reaches $74.97 / 72.26$. The headline gap is largest on $\mathcal{F}_S^{\mu}$ ($+24.04$ absolute), confirming that step-level structured grounding---rather than coarse
answer-region detection---is the central capability HG-Bench measures. Crucially, scale does not close the gap: the $397$B-parameter Qwen3.5-397B-A17B~\cite{bai2025qwen25vl} attains only $42.71 / 18.15$, lower than several smaller closed APIs.

\paragraph{Contributions.}
\begin{itemize}
  \item \textbf{HG-Bench}, the first benchmark for two-level, page-aware answer-region grounding on multi-page handwritten K--12 homework, comprising $500$ human-annotated samples curated from a
$1.49\mathrm{M}$-image source pool.
  \item A \textbf{page-aware evaluation protocol} that decouples the question-level macro metric $\mathcal{F}_A$ from the step-level micro metric $\mathcal{F}_S^{\mu}$ computed over step-bearing pages,
correcting for the structural imbalance of multi-step problems across samples.
  \item A \textbf{systematic evaluation} of nine frontier vision--language systems---closed-source APIs (GPT-5.4, Claude-Sonnet-4.6, Doubao-Seed-2.0-Pro, Gemini-3.0-Pro-Preview) and open-weight models (Qwen3.5-397B-A17B, GLM-5V-Turbo, Kimi-K2.5, GLM-4.6V 9B)~\cite{bai2025qwen25vl,zai2025glmv}---establishing that step-level grounding is the dominant capability gap and that parameter count alone does not close it.
  \item A \textbf{reference fine-tuned system} obtained by single-stage SFT of GLM-4.6V 9B on $\sim$10k in-domain examples, which surpasses every evaluated closed-source baseline without any
reinforcement-learning stage. We release the checkpoint as a lower-bound reference for future HG-Bench submissions.
\end{itemize}

\section{Related Work}
\label{sec:related}

\paragraph{Visual grounding.}
Referring-expression grounding has been studied extensively on natural-image datasets such as the RefCOCO family~\cite{kazemzadeh2014refcoco,yu2016refcocoplus,mao2016refcocog}, Flickr30K
Entities~\cite{plummer2015flickr30k}, and Visual Genome~\cite{krishna2017visualgenome}, which target a single referent per query in editorial photographs. Specialist grounding models such as
GLIP~\cite{li2022glip} and Grounding-DINO~\cite{liu2024groundingdino} push detection-style accuracy on these benchmarks, while recent multimodal LLMs like Shikra~\cite{chen2023shikra} and
Ferret~\cite{you2024ferret} integrate region-level referring into dialogue. HG-Bench instead requires structured multi-region output---a list of per-question regions, each with an ordered list of per-step
sub-regions---on handwritten document scans.


\paragraph{Grounding capabilities of recent VLMs.}
Recent vision--language models---closed-source frontier systems including GPT-4V/4o~\cite{openai2024gpt4v}, Gemini~\cite{google2024gemini}, Claude~\cite{anthropic2024claude},
Doubao~\cite{bytedance2025seed15vl}, and Kimi~\cite{moonshot2026kimik25}, and open-weight families including Qwen-VL / Qwen2.5-VL~\cite{wang2024qwen2vl,bai2025qwen25vl}, InternVL /
InternVL2.5~\cite{chen2024internvl,chen2025internvl25}, CogVLM2~\cite{wang2024cogvlm2}, MiniCPM-V~\cite{yao2024minicpmv}, Florence-2~\cite{xiao2024florence2}, LLaVA-NeXT~\cite{liu2024llavanext},
DeepSeek-VL2~\cite{wu2024deepseekvl2}, Phi-3-Vision~\cite{abdin2024phi3}, and the GLM-V family~\cite{zai2025glmv}---can emit bounding boxes natively. Underlying contrastive backbones such as
CLIP~\cite{radford2021clip} and ALIGN~\cite{jia2021align} supply the visual--linguistic priors these systems inherit, and a growing share of frontier open-weight models adopt Mixture-of-Experts
sparsity~\cite{shazeer2017moe,fedus2022switch,jiang2024mixtral,dai2024deepseekmoe} to scale capacity. Despite this rapid progress, none of these systems has been systematically evaluated on handwritten
K--12 student work, and our results show that none yet solves the task.

\paragraph{Educational AI benchmarks and benchmark methodology.}
Mathematical and scientific reasoning benchmarks such as MathVista~\cite{lu2024mathvista}, MathVerse~\cite{zhang2024mathverse}, We-Math~\cite{qiao2024wemath}, MATH-Vision~\cite{wang2024mathvision}, and
OlympiadBench~\cite{he2024olympiadbench} test problem-solving ability assuming that the relevant content has already been identified. Holistic multimodal evaluation suites such as
MMBench~\cite{liu2024mmbench}, SEED-Bench~\cite{li2023seedbench}, MMMU~\cite{yue2024mmmu}, and the broader HELM~\cite{liang2023helm} and BIG-Bench~\cite{srivastava2023bigbench} programmes target general
capability coverage. None of these efforts measures region-level grounding on handwritten student answers, and benchmark-quality conventions such as Cohen's~$\kappa$~\cite{cohen1960kappa} and
Fleiss'~$\kappa$~\cite{fleiss1971kappa} for inter-annotator agreement remain under-reported in this domain.

\section{Task Formulation}
\label{sec:task}

\paragraph{Problem definition.}
The goal of the task is to evaluate a model's ability to accurately localize student-written answers at both the question and the step level, which is essential for automated grading and trace-of-reasoning
analysis in multi-page handwritten homework.

\paragraph{Inputs and outputs.}
The input to the system is an ordered sequence of homework page images $\{\mathbf{I}_p\}_{p=1}^{P}$ accompanied by metadata specifying each page's pixel dimensions. The expected output is a JSON array
$\{q_i\}_{i=1}^{N}$, where each element $q_i$ corresponds to one question and contains:
\begin{itemize}
    \item a fixed type field \texttt{complete\_answer\_box};
    \item a page index $p_i \in \{1, \dots, P\}$;
    \item a question-level bounding box $\mathbf{b}_i \in [0,1000]^4$ in xyxy format, enclosing the full handwritten answer to the question;
    \item an optional ordered list of step boxes $\{s_{i,j}\}_{j=1}^{K_i}$, each carrying a \texttt{step\_id} (one-indexed in the student's writing order) and its own box $\mathbf{b}_{i,j}$.
\end{itemize}
All question and step boxes must be emitted in the order of the student's answers.

\begin{figure*}[t]
  \centering
  \includegraphics[width=\linewidth]{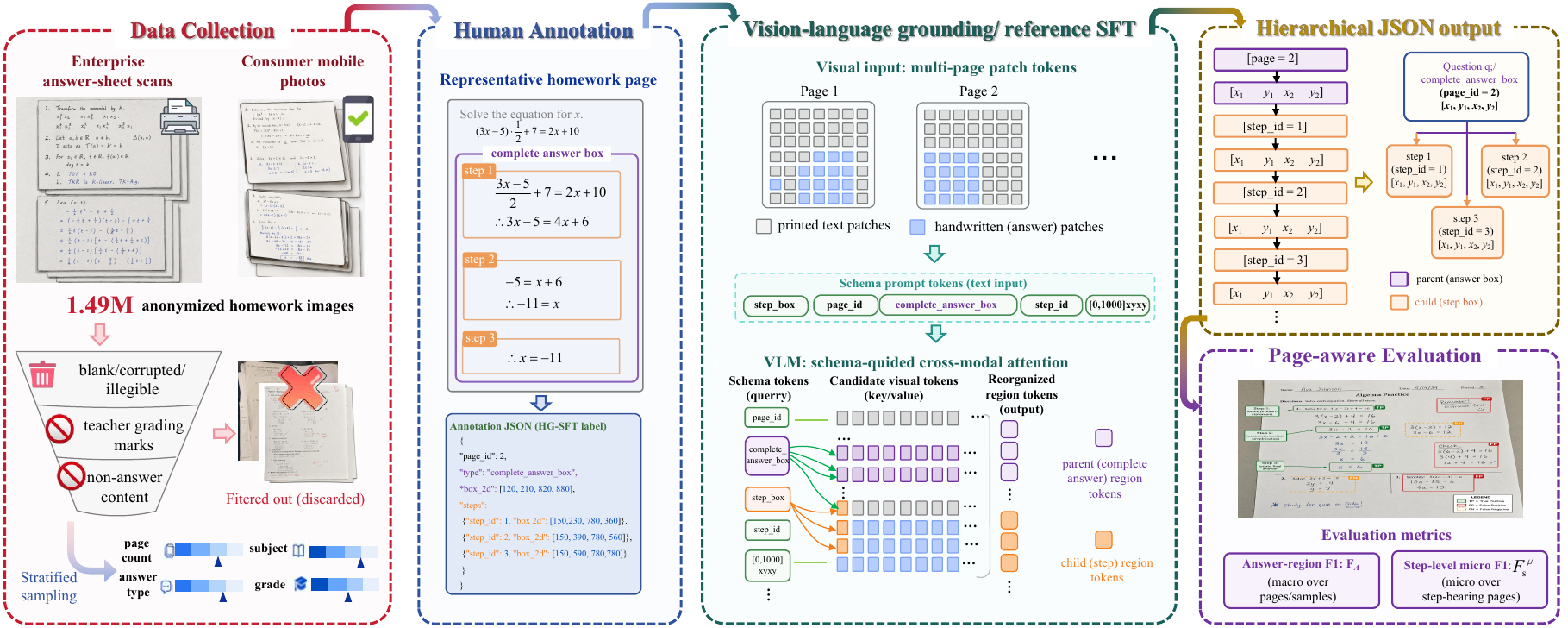}
  \caption{\textbf{HG-Bench data engine.} The pipeline proceeds in four stages. \textbf{(I) Collection.} $1{,}489{,}278$ anonymized homework images are gathered from two complementary channels---enterprise
  B2B answer-sheet scans and consumer photographs from a public-facing consumer homework-photo application---to expose both controlled-capture and in-the-wild distributions. \textbf{(II) Filtering.} Images are filtered for
  usability (corruption, blank, illegibility), for absence of teacher grading marks, and for student-answer content; multi-stage automatic checks are followed by manual verification. \textbf{(III)
Stratified
  annotation.} $10{,}420$ samples are annotated under a two-level boxing protocol (question-level + ordered step-level) with strict hierarchical containment; annotators use a custom tool with
  keyboard-shortcut box drawing. \textbf{(IV) Verification and stratified split.} Every sample is reviewed at least once; ambiguous items are escalated to a lead annotator. The annotated pool is stratified
  along subject, grade, page count, and answer type, yielding a $500$-sample test split (HG-Bench) and a $9{,}920$-sample training pool (HG-SFT) used by the reference system in Sec.~\ref{sec:sft-ref}.}
  \label{fig:pipeline}
\end{figure*}

\paragraph{Two-level box semantics.}
Question-level boxes (occasionally referred to as ``title boxes'' in the annotation tool) localize the complete handwritten answer region of each question and support per-item scoring and partial-credit
attribution. Step-level boxes further decompose multi-step solutions and multi-blank responses into ordered sub-regions, enabling step-level grading and trace-of-reasoning analysis. Each $\mathbf{b}_{i,j}$
must be fully contained within its parent $\mathbf{b}_i$, enforcing hierarchical consistency. When a region is partially missing or ambiguous, models should predict the best-fit bounding box while
preserving this containment rule.

\paragraph{Coordinate convention.}
Coordinates are normalized to the $[0, 1000]$ xyxy format by default and can be denormalized to pixel values using per-page metadata. The protocol also supports yxyx and pixel-coordinate variants via
configuration. Models that emit polygons are evaluated via the minimum enclosing axis-aligned rectangle. All coordinates must tightly enclose student-written content and exclude printed text and teacher
annotations.


\section{HG-Bench}
\label{sec:bench}

\subsection{Source Pool}
HG-Bench is derived from a large pool of $1{,}489{,}278$ anonymized student homework images, comprising real online homework scans and ink-screen captures spanning multiple subjects and grade levels. From
this raw pool, we curate a high-quality annotated set of $10{,}420$ valid samples from two representative collection channels:
\begin{itemize}
    \item \textbf{Enterprise channel} ($6{,}765$ samples), drawn from formal answer sheets and standardized exam grading, characterized by multi-page samples and a high proportion of multi-step problems;
    \item \textbf{Consumer channel} ($3{,}655$ samples), drawn from homework photos uploaded by general users of a public-facing consumer homework-photo application, typically single-page images with a more balanced distribution
of question types and in-the-wild capture variation.
\end{itemize}
The annotated pool is then partitioned into a $500$-sample held-out test set (\textbf{HG-Bench}) and a $9{,}920$-sample training pool (\textbf{HG-SFT}), with $250$ samples held out from each channel. The
test and training pools are disjoint by construction; we additionally verify disjointness with perceptual hashing (Sec.~\ref{sec:sft-ref}). All personally identifying information was removed before
inclusion in either pool.

\subsection{Sampling Strategy}
To ensure that the benchmark is representative of real-world homework, we perform stratified sampling along four axes: subject, grade level, page count per sample, and answer type. This procedure yields the
 $500$ benchmark samples and is designed to capture the long tail of layouts and problem complexities encountered in practice. Detailed per-stratum counts appear in Table~\ref{tab:stats} and in
Figure~\ref{fig:data_distributions}.

\subsection{Annotation Protocol}
The annotation protocol specifies which pages to skip and prescribes the procedure for drawing the two levels of bounding boxes.

\paragraph{Skip rules.}
A page is skipped if (i) the image is unusable (corrupted, blank, or illegible); (ii) teacher grading marks (\checkmark, $\times$, written scores, etc.) are present; or (iii) all student answers are
non-conventional (drawings, doodles, off-task content). Pages containing red-pen marks not associated with grading are retained and annotated normally.

\paragraph{Box drawing.}
Each region of student handwriting is enclosed by an axis-aligned bounding box. Single-answer questions receive one question-level box. Multi-step solutions and multi-blank responses additionally receive an
 ordered set of step-level boxes, each tightly enclosing one step's handwriting and never splitting a single handwritten line across multiple boxes. Informal scratch work is not boxed.

\paragraph{Tagging.}
Each box carries the local question number and the parent-number hierarchy, separated by ``/''. Question type is tagged from a fixed inventory: choice, fill, judgment, solve (including computation),
drawing, short-answer, writing. Step IDs are integers assigned in the order of the student's writing. All punctuation in tags follows Chinese typographic conventions.

\subsection{Annotation Workflow}
Annotations followed a two-stage protocol. 
\textbf{12} trained annotators drew question- and step-level boxes with a custom shortcut-based tool, and \textbf{5} senior reviewers independently accepted each annotation or returned it for revision; ambiguous cases were escalated to a lead annotator. 
We measured inter-annotator agreement (IAA) on $50$ randomly sampled test samples annotated by two independent annotators before QC revisions. 
For localization, the annotations achieved a mean IoU of $0.86$, with $85\%$ of boxes matched at $\mathrm{IoU}=0.5$, and Cohen's $\kappa=0.83$ for binary box-matching consistency. 
For discrete annotations---question type, step containment, and step ordering---Fleiss' $\kappa$ was $0.81$. 
These results indicate substantial agreement~\citep{cohen1960kappa,fleiss1971kappa} and support the reliability of HG-Bench.

\subsection{Dataset Statistics}
\begin{figure}[t]
\centering
\includegraphics[width=\linewidth]{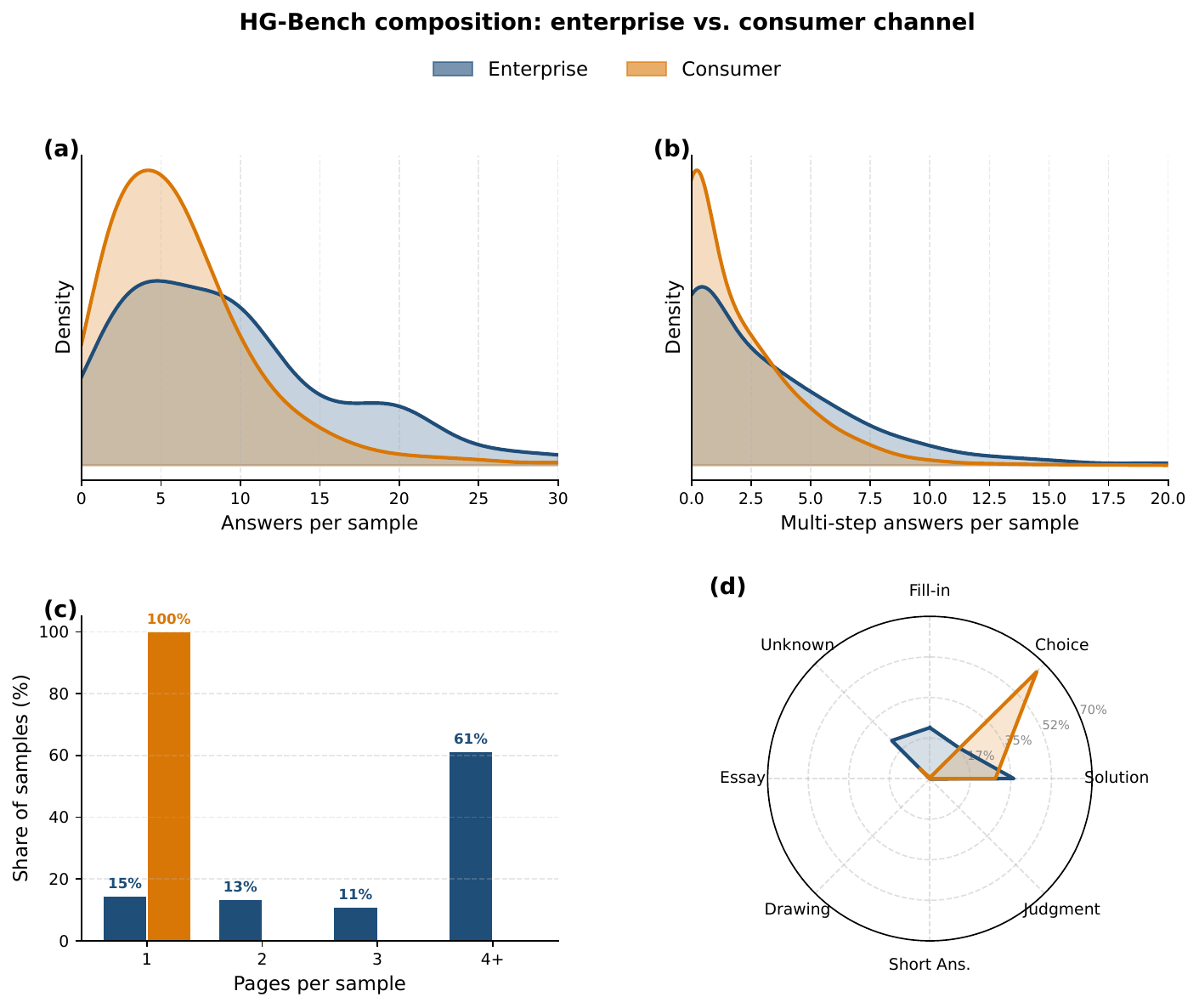}
\caption{\textbf{HG-Bench composition.} (a) Answers per sample follow a long-tail distribution in both enterprise and consumer channels. (b) Multi-step problems are over-represented in the enterprise
channel, consistent with formal exam scenarios. (c) Page count per sample: enterprise samples are predominantly multi-page; consumer samples are predominantly single-page. (d) Question-type composition:
fill-in-the-blank dominates the enterprise channel, while the consumer channel is more uniformly distributed across choice, fill, solve, and short-answer types. The $500$-sample benchmark is stratified
along these four axes to remain representative of both channels.}
\label{fig:data_distributions}
\end{figure}

We summarize the annotated pool and HG-Bench test set in Figure~\ref{fig:data_distributions} and Table~\ref{tab:stats}. 
The data reflect realistic homework scenarios: both channels show long-tailed answer counts, while enterprise samples contain more multi-step problems, more multi-page submissions, and more fill-in-the-blank items. 
Consumer samples are mostly single-page and more balanced across choice, fill-in-the-blank, solve, and short-answer questions. 
The $500$ HG-Bench samples are stratified to preserve these patterns, making the benchmark representative and challenging for both question-level and step-level grounding.

\begin{table}[t]
  \centering
  \caption{HG-Bench summary statistics.}
  \label{tab:stats}
  \footnotesize
  \begin{tabular}{lr}
    \toprule
    \textbf{Statistic} & \textbf{Value} \\
    \midrule
    Samples                                & $500$ \\
    Subjects                               & $5$ \\
    Mean pages per sample                  & $1.8$ \\
    Mean question boxes per page           & $5.2$ \\
    Mean step boxes per question (when present) & $2.9$ \\
    Fraction of step-bearing pages         & $0.34$ \\
    \bottomrule
  \end{tabular}
\end{table}


\section{Evaluation Protocol}
\label{sec:eval}


\subsection{Page-Aware Box Matching}
Unlike standard grounding tasks that flatten all bounding boxes across an entire document, our evaluation operates strictly at the page level to preserve the multi-page layout structure.

Within each page, we perform greedy one-to-one matching between ground-truth boxes $\mathcal{G}$ and predicted boxes $\mathcal{P}$ under an Intersection-over-Union (IoU) constraint. Specifically, we filter
out all candidate pairs $(g, p) \in \mathcal{G} \times \mathcal{P}$ with $\mathrm{IoU}(g,p) < 0.5$. The remaining pairs are sorted in descending order of IoU and assigned greedily, so that each box
participates in at most one match. This localized matching yields per-page true-positive ($\mathrm{TP}$), false-positive ($\mathrm{FP}$), and false-negative ($\mathrm{FN}$) counts.

\subsection{Reported Metrics}
Following page-level matching, performance is aggregated into two primary dataset-level localization metrics; we additionally report four supplementary metrics defined in Appendix~\ref{sec:appendix_metrics} to characterize step-decomposition robustness ($\mathcal{F}_S^{M}$) and parse-time reliability ($\mathrm{Succ}\%$, $\bar{\mathcal{S}}$, $\mathrm{Rep}\%$).

\begin{itemize}
    \item \textbf{Answer-region $\mathrm{F}_1$ ($\mathcal{F}_A$).} This metric evaluates the localization quality of question-level answer regions. We compute the standard $\mathrm{F}_1$ score per page from
 its local $\mathrm{TP}$, $\mathrm{FP}$, and $\mathrm{FN}$ counts, average within each multi-page sample, and report the macro-average across all successfully evaluated samples:
\begin{equation}
    \resizebox{1\linewidth}{!}{%
        $\mathcal{F}_A = \frac{1}{|\mathcal{S}_{\text{succ}}|} \sum_{s \in \mathcal{S}_{\text{succ}}} \left( \frac{1}{|\mathcal{M}_s|} \sum_{m \in \mathcal{M}_s} \mathrm{F}_1(s, m) \right) \times 100$%
    }
\end{equation}
    where $\mathcal{S}_{\text{succ}}$ is the set of successfully evaluated samples, $\mathcal{M}_s$ is the set of pages within sample $s$, and $\mathrm{F}_1(s, m)$ is the box-level $\mathrm{F}_1$ on the
$m$-th page of sample $s$.

    \item \textbf{Step-level micro $\mathrm{F}_1$ ($\mathcal{F}_{S}^{\mu}$).} Unlike question-level answer regions, which are present on every page, step-level boxes exhibit a highly sparse and imbalanced
distribution: many pages contain no steps at all. To eliminate the evaluation bias induced by this imbalance, $\mathcal{F}_S^{\mu}$ is computed by micro-aggregation restricted to step-bearing pages. We
accumulate step-level $\mathrm{TP}$, $\mathrm{FP}$, and $\mathrm{FN}$ counts globally across all pages whose ground truth contains at least one step box and compute a single unified $\mathrm{F}_1$ score.
\end{itemize}

\section{Benchmarking and Experimental Analysis}
\label{sec:bench_results}

All baseline VLMs use the same prompt template (Appendix~\ref{sec:appendix_prompts}), which fixes the JSON schema, normalized $[0,1000]$ coordinates, and page-aware sequence-preserving output format. 
We parse outputs with a format-tolerant JSON parser and issue one format-reminder retry after structural failures; persistent failures are marked with \texttt{FAIL\_STR} ($\texttt{success}=\textit{False}$). 
Because models differ in coordinate conventions, we auto-detect box-axis ordering per model on a small held-out calibration slice before evaluation.

\subsection{Evaluated Baselines}
We evaluate two cohorts of frontier vision--language systems:
\begin{itemize}
    \item \textbf{Closed-source frontier APIs.} GPT-5.4~\cite{openai2026gpt54}, Claude-Sonnet-4.6~\cite{anthropic2026sonnet46}, Doubao-Seed-2.0-Pro (snapshots 2026-02-15 and 2026-04-01), and Gemini-3.0-Pro-Preview~\cite{google2024gemini}, called through their official endpoints under default decoding.
    \item \textbf{Open-weight baselines.} Qwen3.5-397B-A17B~\cite{bai2025qwen25vl} (a Mixture-of-Experts model activating $17$B parameters per token), GLM-5V-Turbo, Kimi K2.5, and the GLM-4.6V 9B base~\cite{zai2025glmv}.
\end{itemize}
We additionally report \textbf{GLM-4.6V-9B + HG-SFT}, a reference fine-tuned system trained on the HG-SFT pool. 
This model is included as a trained reference point rather than a zero-shot baseline, allowing us to test whether HG-Bench is learnable with a modest amount of in-domain supervision.

\subsection{Reference SFT System}
\label{sec:sft-ref}

To verify learnability and provide a reproducible lower-bound reference, we fine-tune GLM-4.6V 9B on the $9{,}920$-sample HG-SFT training pool using single-stage supervised fine-tuning. 
The training uses no reinforcement learning, no synthetic continued pre-training, and no out-of-domain data mixing, so improvements over zero-shot baselines can be attributed to targeted in-domain supervision rather than to a stronger foundation model or additional training stages. 
Full details on the base checkpoint, train--test deduplication, data composition, and optimization recipe are provided in Appendix~\ref{app:sft-ref-details}.

\subsection{Main Results}
\label{sec:main-results}
Table~\ref{tab:main} reports results across all baselines and the reference SFT system.


\begin{table*}[t]
\centering
\caption{
Main results on HG-Bench ($N=500$ samples for every model).
$\mathcal{F}_A$: macro answer-region $\mathrm{F}_1$ averaged across samples and pages.
$\mathcal{F}_S^{\mu}$: micro step-level $\mathrm{F}_1$ aggregated over step-bearing pages.
$\mathcal{F}_S^{M}$: macro step-level $\mathrm{F}_1$ averaged over step-bearing samples (Appendix~\ref{sec:appendix_metrics}).
$\mathrm{Succ}\%$: parse success rate.
$\bar{\mathcal{S}}$: unified composite score averaged over all $500$ samples (failed parses count as $0$).
$\mathrm{Rep}\%$: fraction of outputs containing repeated content (lower is better).
Best results are shown in bold, second-best underlined.
}
\label{tab:main}
\setlength{\tabcolsep}{5.0pt}
\renewcommand{\arraystretch}{1.1}
\resizebox{0.7\textwidth}{!}{
\begin{tabular}{@{}l||cccccc@{}}
\toprule[1.2pt]
\rowcolor{headergray}
\textbf{Model}
& $\boldsymbol{\mathcal{F}_A}$
& $\boldsymbol{\mathcal{F}_S^{\mu}}$
& $\boldsymbol{\mathcal{F}_S^{M}}$
& \textbf{Succ\%}
& $\boldsymbol{\bar{\mathcal{S}}}$
& \textbf{Rep\%} \\
\midrule[1.0pt]

\multicolumn{7}{@{}l}{\textit{\textcolor{gray}{Closed-source frontier APIs}}} \\
\rowcolor{rowgray}
GPT-5.4
& 14.91 & 1.55 & 1.38 & 100.0 & 8.12 & 0.4 \\
Claude-Sonnet-4.6
& 16.83 & 1.63 & 1.21 & 99.2 & 8.76 & 2.8 \\
\rowcolor{rowgray}
Doubao-Seed-2.0-Pro (2026-02-15)
& 52.65 & 44.78 & \second{42.59} & 49.4 & 21.22 & \best{0.0} \\
Doubao-Seed-2.0-Pro (2026-04-01)
& \second{55.22} & 40.11 & 34.88 & 99.8 & \second{42.70} & \second{0.2} \\
\rowcolor{rowgray}
Gemini-3.0-Pro-Preview
& 50.90 & \second{48.22} & 37.58 & 100.0 & 42.33 & 6.0 \\

\midrule
\multicolumn{7}{@{}l}{\textit{\textcolor{gray}{Open-weight baselines}}} \\
\rowcolor{rowgray}
Qwen3.5-397B-A17B
& 42.71 & 18.15 & 17.73 & 94.2 & 32.73 & 4.0 \\
GLM-5V-Turbo
& 46.69 & 26.29 & 23.78 & 100.0 & 40.10 & 0.4 \\
\rowcolor{rowgray}
Kimi K2.5
& 31.21 & 7.42 & 7.21 & 79.4 & 20.18 & 1.0 \\
GLM-4.6V 9B (base)
& 34.15 & 7.65 & 4.60 & 100.0 & 29.46 & 3.8 \\

\midrule
\multicolumn{7}{@{}l}{\textit{\textcolor{gray}{Reference SFT system}}} \\
\rowcolor{oursblue}
\textbf{GLM-4.6V-9B + HG-SFT}
& \best{74.97} & \best{72.26} & \best{48.25} & 100.0 & \best{71.53} & 0.8 \\
\rowcolor{oursblue}
\textcolor{gainblue}{\small $\Delta$ over best prior}
& \gain{19.75} & \gain{24.04} & \gain{5.66} & -- & \gain{28.83} & -- \\

\bottomrule[1.2pt]
\end{tabular}
}
\end{table*}

\noindent\textbf{Frontier VLMs plateau well below a trained reference, and scale alone does not close the gap.}
No zero-shot baseline---closed- or open-source---exceeds $55.22$ on $\mathcal{F}_A$ or $48.22$ on $\mathcal{F}_S^{\mu}$. The $\sim$10k-example reference system reaches $74.97 / 72.26$, leaving headroom of
roughly $20$ and $24$ absolute points over the best zero-shot result on each metric. The gap is not explained by parameter count: the $397$B-parameter Qwen3.5-397B-A17B remains at $42.71 / 18.15$, weaker
than several smaller closed APIs on $\mathcal{F}_A$ and below the consumer-class GLM-5V-Turbo on $\mathcal{F}_S^{\mu}$. We read this as evidence that HG-Bench measures a capability axis---fine-grained,
ordered cross-modal localization---that is not addressed by general-purpose pre-training scale, and that the benchmark is far from saturated.

\noindent\textbf{Step-level grounding is the dominant capability gap.}
All zero-shot baselines degrade sharply from $\mathcal{F}_A$ to $\mathcal{F}_S^{\mu}$, showing that locating complete answer regions is much easier than decomposing ordered reasoning steps. 
Open-weight models such as Kimi K2.5 and GLM-4.6V 9B fall to single-digit step scores, while GPT-5.4 and Claude-Sonnet-4.6 nearly collapse ($1.55$ and $1.63$); even the strongest closed API drops from $55.22$ to $40.11$. 
In contrast, the reference SFT system reduces this gap to about $3$ points, suggesting that step-level grounding benefits from targeted supervision rather than scale alone. 
So we rank HG-Bench primarily by $\mathcal{F}_S^{\mu}$.


\begin{figure*}[t]
  \centering
  \includegraphics[width=1\textwidth]{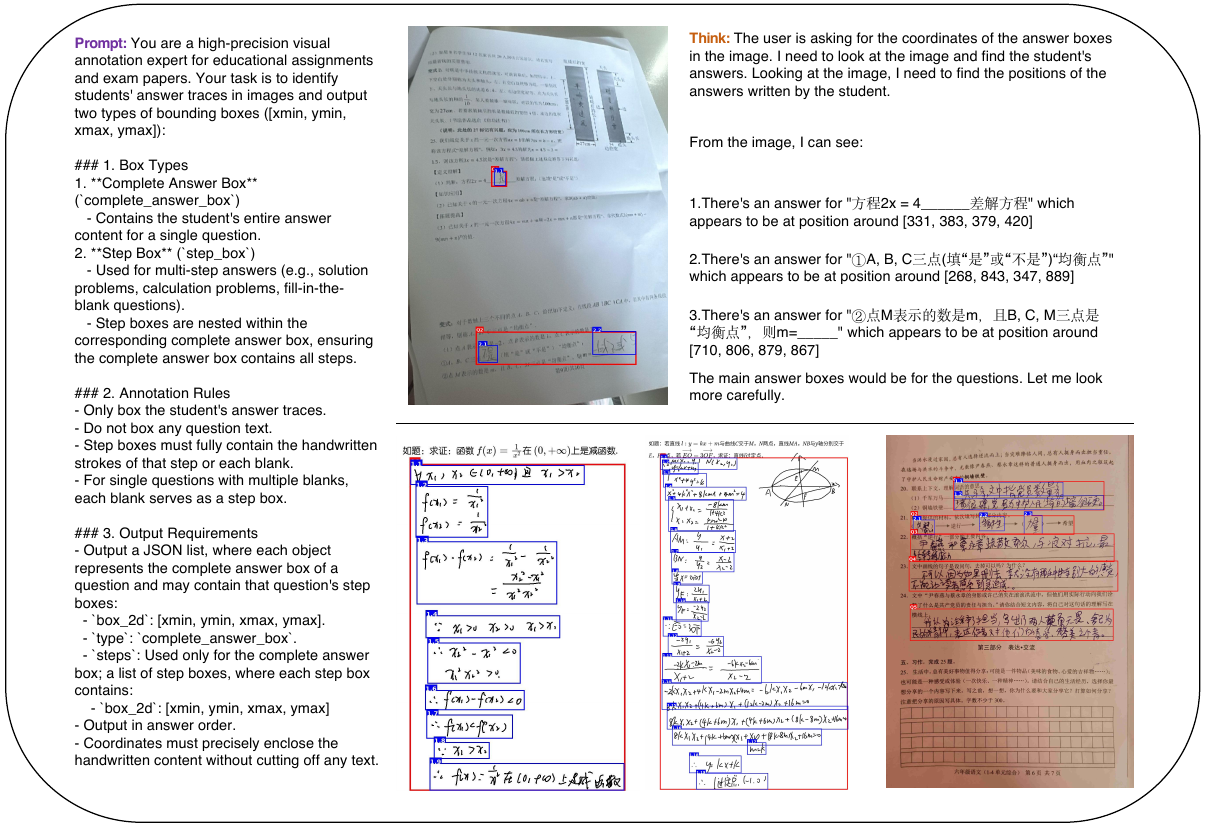}
  \caption{\textbf{Task overview and sample diversity on HG-Bench.}
  \emph{Top}: the shared evaluation prompt, a representative two-page handwritten input with predicted question-level boxes (red) and step-level boxes (blue), and the model's textual grounding trace.
  \emph{Bottom}: three additional samples spanning function derivation, calculation, and Chinese long-form writing, showing the benchmark's diversity in layout, handwriting density, and subject type.
  Additional failure-specific cases appear in Appendix~\ref{sec:appendix_cases}.}
  \label{fig:qual}
\end{figure*}

\subsection{Subject and Page-Count Breakdown}
\label{sec:bench-breakdown}

\vspace{1mm}
\noindent\textbf{Subject breakdown.}
Across the evaluated subjects, zero-shot baselines score lowest on Mathematics and Science. We attribute this to the non-linear layout of STEM solutions---fraction derivations, matrices, spatial scratch
notes---which violates the predominantly left-to-right, top-to-bottom prior of generic VLM training. Language subjects (Chinese, English) follow a more regular linear sequence, making question-level
bounding ($\mathcal{F}_A$) easier; even so, fine-grained step tracking ($\mathcal{F}_S^{\mu}$) degrades on long-form answers with dense interlocking paragraphs, indicating that the bottleneck is layout
density rather than subject identity per se.

\vspace{1mm}
\noindent\textbf{Page-count scaling.}
We slice the test set by input page count ($1$, $2$, $3+$). Zero-shot baselines retain coordinate formatting on single-page inputs but exhibit monotonic decay in step-level metrics as page count grows;
across baselines, $\mathcal{F}_S^{\mu}$ drops by over $25\%$ on average from $1$ to $3+$ pages. The failure mode is consistent across families: page-boundary hallucinations (boxes attributed to the wrong
page) and index shuffling (out-of-order step IDs across page breaks). The reference SFT system shows substantially smaller decay, indicating that exposure to multi-page context during SFT stabilizes
cross-page coordinate mappings. We flag multi-page coordinate stability as an open problem that current cross-modal attention does not solve from scale alone.

\section{Analysis}
\label{sec:analysis}


We categorize model failures into five types: \textbf{hallucinated boxes} covering no handwriting, \textbf{missed steps} where a multi-step answer receives only a question-level box, \textbf{page misalignment} with boxes assigned to the wrong page, \textbf{step over-/under-segmentation} where one step is split or multiple steps are merged, and \textbf{format failures} such as invalid JSON or schema violations. 
Closed-source frontier systems mainly fail through missed steps and under-segmentation, while weaker open-weight models additionally suffer from frequent format errors. 
Figure~\ref{fig:qual} illustrates the task setting and representative HG-Bench samples; detailed failure cases are provided in Appendix~\ref{sec:appendix_cases}.

\subsection{Inter-Model Agreement}
\label{sec:inter-model}

To measure how well HG-Bench separates model capabilities, we count, for each test sample, how many of the nine zero-shot baselines achieve question-level $\mathrm{F}_1 \geq T$, using $T=0.5$ as the standard matching threshold and $T=0.7$ as a stricter high-quality bar.

\vspace{1mm}
\noindent\textbf{HG-Bench is far from saturated.}
At $T=0.5$, only $3/500$ samples ($0.6\%$) are solved by all nine baselines, while $115$ samples ($23.0\%$) are missed by every baseline; the remaining $76\%$ are solved by one to eight models, indicating a smooth difficulty gradient. 
At $T=0.7$, the universally missed set rises to $44.6\%$, whereas the all-passed set remains $0.6\%$, confirming that current frontier VLMs leave substantial headroom.

\vspace{1mm}
\noindent\textbf{Hard samples are not dominated by multi-page inputs.}
Among the $115$ universally missed samples at $T=0.5$, $59$ ($51\%$) come from the enterprise channel and $56$ ($49\%$) from the consumer channel, closely matching the balanced test-set composition. 
Thus, title-level failures are not explained by page count alone; handwriting density and step decomposition remain the main residual challenges.

\vspace{1mm}
\noindent\textbf{Targeted SFT recovers genuinely hard cases.}
On the $115$ samples missed by every zero-shot baseline at $T=0.5$, the reference SFT system reaches question-level $\mathrm{F}_1 \geq 0.5$ on $68$ samples ($59.1\%$). 
Under the stricter $T=0.7$ threshold, it still rescues $45.7\%$ of the $223$ universally missed samples. 
This shows that HG-SFT does not merely improve easy cases, but specifically recovers examples beyond the reach of frontier zero-shot systems, supporting the learnability claim in Sec.~\ref{sec:main-results}.

\section{Conclusion}
\label{sec:conclusion}

We introduced HG-Bench, the first benchmark for per-question and per-step answer-region grounding on multi-page handwritten K--12 homework, with a page-aware protocol for imbalanced step-bearing pages. 
Evaluating frontier closed-source and open-weight VLMs shows that $\mathcal{F}_A$ plateaus around $50$--$55$, while step-level grounding ($\mathcal{F}_S^{\mu}$) remains the main bottleneck, with several models falling below $10$. 
A simple SFT reference based on an open-weight $9$B model and $\sim$10k in-domain examples surpasses all evaluated closed-source systems without reinforcement learning, highlighting targeted supervision as a practical path toward structured homework grounding.

\section*{Limitations}

HG-Bench targets Chinese K--12 homework; transfer to other languages
and to higher-education work is not measured. The benchmark contains
$500$ samples, which limits the resolution of finer-grained subject-
or grade-level conclusions. Evaluation uses a single $\mathrm{IoU}\!=\!0.5$
matching threshold; performance under tighter or looser thresholds
is not reported here. Step decomposition contains an irreducible
subjective component, particularly for short solve-type answers,
which the two-stage annotation protocol (Section~\ref{sec:bench})
mitigates but does not eliminate.
\textbf{Inter-annotator agreement} is reported on a $50$-sample
randomly sampled subset with two independent annotators
(Section~\ref{sec:bench}); extending IAA to a larger subset and
reporting per-category and question/step breakdowns is a near-term
release item.
\textbf{Reference-system ablations} (channel ablation, data scaling,
training-length scaling) are likewise deferred to future work building
on the released HG-SFT corpus; the reference SFT system is reported
as a learnability lower bound rather than as a methods contribution.
Finally, the reference system was trained without a reinforcement-learning
stage; whether RL would further improve performance on HG-Bench

\section*{Ethics Statement}

Source homework images were anonymized prior to inclusion in either the benchmark or the training pool. No personally identifying information about individual students appears in either resource, and no
per-student attribute is exposed to the evaluated models. The intended downstream use of systems built on HG-Bench is teacher-side grading assistance, not automated student evaluation, ranking, or
admissions decisions. We do not release individual student work; benchmark images are obtained through a commercial partnership with a third-party data provider under terms permitting educational research use, and we redistribute only derived annotation metadata together with anonymized identifiers.

\clearpage

\appendix

\section{Annotation Guidelines (Excerpts)}
\label{sec:appendix_guidelines}

This appendix summarizes the core rules used by the annotator pool
(Section~\ref{sec:bench}). The full guideline document is released
with the benchmark.

\paragraph{Skip rules.} A page is \emph{skipped} (no boxes drawn)
when any of the following apply: (i) the image is corrupted, blank,
or illegible; (ii) teacher grading marks are present anywhere on the
page (check marks, crosses, written scores, marker corrections); or
(iii) all student-produced marks on the page are non-conventional
(free drawings, doodles, off-task content). Red-pen marks that are
clearly part of the student's own answer (e.g., underlining or
highlighting) are not treated as grading marks and the page is
retained.

\paragraph{Two-level boxing.} Every region of student handwriting is
enclosed by an axis-aligned bounding box. Each question receives
exactly one question-level \texttt{complete\_answer\_box} that tightly
bounds all of the student's handwriting attributable to that
question. For multi-step solutions (computation, derivation,
multi-blank fill-in), an additional ordered list of \texttt{step\_box}
elements decomposes the answer; each step box must be fully contained
within its parent question-level box. A single handwritten line may
never be split across two boxes. Informal scratch work (calculations
in the margin, crossed-out drafts) is not boxed.

\paragraph{Choice and judgment answers.} When both a bubble-fill
region and a hand-written letter answer are present, annotators box
the bubble region. When only one is present, that one is boxed.

\paragraph{Tagging.} Each question-level box carries (i) the parent
title number and (ii) the sub-question number, separated by a slash.
The original tags follow Chinese typographic convention (full-width
parentheses, circled digits, blank-N identifiers); the released
schema preserves the original strings verbatim. Step IDs are
integers assigned in the order in which the student wrote the steps,
starting from 1.

\paragraph{Hierarchical containment.} The containment property---each
step box contained in its parent question-level box---is enforced at
annotation time and re-verified by the QC reviewer.

\section{Reference SFT System Details}
\label{app:sft-ref-details}

This appendix provides the full configuration of the reference SFT system reported in Section~\ref{sec:sft-ref}. 
The system is intended as a learnability probe and reproducible lower-bound reference for HG-Bench, not as a state-of-the-art model submission.

\paragraph{Base checkpoint.}
We initialize from GLM-4.6V 9B~\cite{zai2025glmv}, an open-weight vision--language model released by Z.ai on 2025-09-30. 
We deliberately avoid initializing from any newer same-family checkpoint, ensuring that gains over the baselines in Table~\ref{tab:main} cannot be explained by a stronger or more recent foundation model.

\paragraph{Training data.}
We fine-tune on \textbf{HG-SFT}, the $9{,}920$-sample training pool derived from the same annotation effort as the $500$-sample HG-Bench test set. 
The two pools are disjoint by construction. 
As an additional safeguard, every training image is checked against the HG-Bench test pool using perceptual hashing (pHash, Hamming distance $\le 5$) together with exact metadata matching on user ID and capture timestamp. 
Before deduplication, the raw candidate training pool contained $14{,}264$ samples; pHash-based filtering removed $4{,}344$ near-duplicates, yielding the final $N_{\text{train}}=9{,}920$. 
HG-SFT preserves the natural enterprise / consumer composition of the source pool ($6{,}515 / 3{,}405$), and we do not re-weight between channels.

\paragraph{Optimization recipe.}
We perform single-stage supervised fine-tuning for $3$ epochs, corresponding to $930$ optimization steps, with global batch size $32$, sequence length $32{,}768$, bfloat16 precision, and AdamW optimization. 
The learning rate follows a cosine schedule with $30$ warmup steps, peak learning rate $1\!\times\!10^{-6}$, minimum learning rate $5\!\times\!10^{-7}$, weight decay $0.1$, and gradient clipping at $1.0$. 
Training runs on a single $8\!\times\!\text{H100-80GB}$ node with tensor parallelism $=2$, context parallelism $=2$, and sequence parallelism enabled, including the ViT tower. 
Table~\ref{tab:sft-hp} lists the complete hyperparameter and system configuration.

\begin{table}[h]
\centering
\small
\resizebox{\linewidth}{!}{%
\begin{tabular}{ll}
\toprule
\textbf{Item} & \textbf{Value} \\
\midrule
Base checkpoint & GLM-4.6V 9B~\cite{zai2025glmv} \\
Precision & bfloat16 \\
Sequence length & $32{,}768$ \\
Visual tokens / image & up to $10{,}000$ (variable shape, jitter $0.75$--$1.25$) \\
Optimizer & AdamW, $(\beta_1, \beta_2)\!=\!(0.9, 0.95)$, $\epsilon\!=\!10^{-8}$ \\
Weight decay & $0.1$ \\
Gradient clipping & $1.0$ \\
LR schedule & cosine, $30$-step warmup \\
Peak / min LR & $1\!\times\!10^{-6}$ / $5\!\times\!10^{-7}$ \\
Global batch size & $32$ (micro-batch $1$) \\
Training steps & $930$ ($=3$ epochs over $9{,}920$ examples) \\
Dropout & $0$ \\
\midrule
Hardware & $1$ node, $8\!\times\!\text{H100-80GB}$ \\
Tensor parallel & $2$ \\
Context parallel & $2$ \\
Sequence parallel & enabled, including ViT tower \\
Pipeline parallel & $1$ \\
ZeRO / distributed optimizer & enabled, overlapped gradient reduction \\
Activation recomputation & full, per block \\
Activation offload & enabled, including ViT convolution and projection \\
Optimizer-state offload & CPU, $100\%$, precision-aware moments \\
Wall-clock time & $\approx 3$ hours \\
\midrule
Total compute & $\approx 24$ GPU-hours \\
\bottomrule
\end{tabular}%
}
\caption{Hyperparameters and system configuration of the reference SFT system.}
\label{tab:sft-hp}
\end{table}

\section{Prompt Templates}
\label{sec:appendix_prompts}

\paragraph{Evaluation prompt.}
All baseline VLMs are queried with a single unified prompt template
(English translation below; the verbatim Chinese version actually
used in evaluation is released with the benchmark).

\begin{quote}\small
You are a high-precision visual annotation expert for educational
homework and exam papers. Your task is to identify each student's
handwritten answer regions in the provided images and output two
types of bounding boxes, with coordinates in
\texttt{[xmin, ymin, xmax, ymax]} format normalized to
\texttt{[0, 1000]}.

\textbf{Box types.}
(a)~\texttt{complete\_answer\_box}: tightly contains the student's
entire answer to one question. For multiple-choice and
true/false items, if both a bubble-fill region and a hand-written
letter answer are present, prefer the bubble-fill region.
(b)~\texttt{step\_box}: used for multi-step answers such as
computation, derivation, and multi-blank fill-in items. Each step or
blank is boxed separately and assigned a \texttt{step\_id} starting
from 1 in the order the student wrote them. Every step box must be
nested inside the corresponding \texttt{complete\_answer\_box}.

\textbf{Annotation rules.} Box only the student's own marks
(handwritten text, edits, ticks, connecting lines, drawings). Do not
box printed question text or teacher corrections. Step boxes must
fully contain the handwritten content of that step or blank. For a
single multi-blank item, each blank becomes a separate step box in
left-to-right, top-to-bottom order.

\textbf{Output format.} Emit a JSON list. Each element is one
question-level object with the following fields:
\texttt{box\_2d} (the question-level box),
\texttt{type} (fixed to \texttt{complete\_answer\_box}),
and an optional \texttt{steps} list whose elements each carry their
own \texttt{box\_2d} and integer \texttt{step\_id}.
Items must be emitted in the order the student answered.
Coordinates must tightly bound the handwriting without cropping any
character.
\end{quote}
A schematic example of the expected JSON output is shown in
Figure~\ref{fig:json-example}.

\begin{figure}[ht]
\centering
\begin{tcblisting}{
  width=0.96\linewidth,
  colback=jsonbg,
  colframe=jsonframe,
  coltext=jsontext,
  boxrule=0.6pt,
  arc=2mm,
  left=2mm,
  right=2mm,
  top=1mm,
  bottom=1mm,
  enhanced,
  listing only,
  listing options={
    basicstyle=\ttfamily\scriptsize,
    columns=fullflexible,
    breaklines=true,
    keepspaces=true,
    showstringspaces=false
  }
}
[
  {
    "box_2d": [100, 200, 180, 300],
    "type": "complete_answer_box"
  },
  {
    "box_2d": [400, 220, 490, 320],
    "type": "complete_answer_box",
    "steps": [
      {"box_2d": [410, 230, 440, 320], "step_id": 1},
      {"box_2d": [450, 230, 480, 320], "step_id": 2}
    ]
  },
  {
    "box_2d": [500, 220, 580, 780],
    "type": "complete_answer_box",
    "steps": [
      {"box_2d": [510, 230, 540, 780], "step_id": 1},
      {"box_2d": [550, 230, 580, 780], "step_id": 2},
      {"box_2d": [590, 230, 620, 780], "step_id": 3}
    ]
  }
]
\end{tcblisting}
\caption{Example JSON output illustrating the two-level box schema: one multiple-choice question with no steps, one fill-in item with two ordered blanks, and one solve item with three ordered derivation steps.}
\label{fig:json-example}
\end{figure}

\paragraph{Format-reminder retry prompt.} When the format-tolerant
parser fails to recover a valid JSON array from a model's first
reply, a single retry is issued with the appended instruction:
\begin{quote}\small\itshape
Your previous response could not be parsed as a valid JSON array.
Please reply with only the JSON array as described above, with no
surrounding prose or markdown code fences.
\end{quote}
Persistent failures after this retry are recorded as terminal errors
via the sentinel \texttt{FAIL\_STR}
($\texttt{success}\!=\!\textit{False}$).

\section{Supplementary Metrics}
\label{sec:appendix_metrics}

In addition to the two primary localization metrics ($\mathcal{F}_A$ and $\mathcal{F}_S^{\mu}$) reported in Section~\ref{sec:eval}, we record four supplementary metrics in Table~\ref{tab:main} to characterize the macro behavior of step decomposition and the parse-time reliability of each VLM.

\begin{figure*}[ht]
  \centering
  \includegraphics[width=\linewidth]{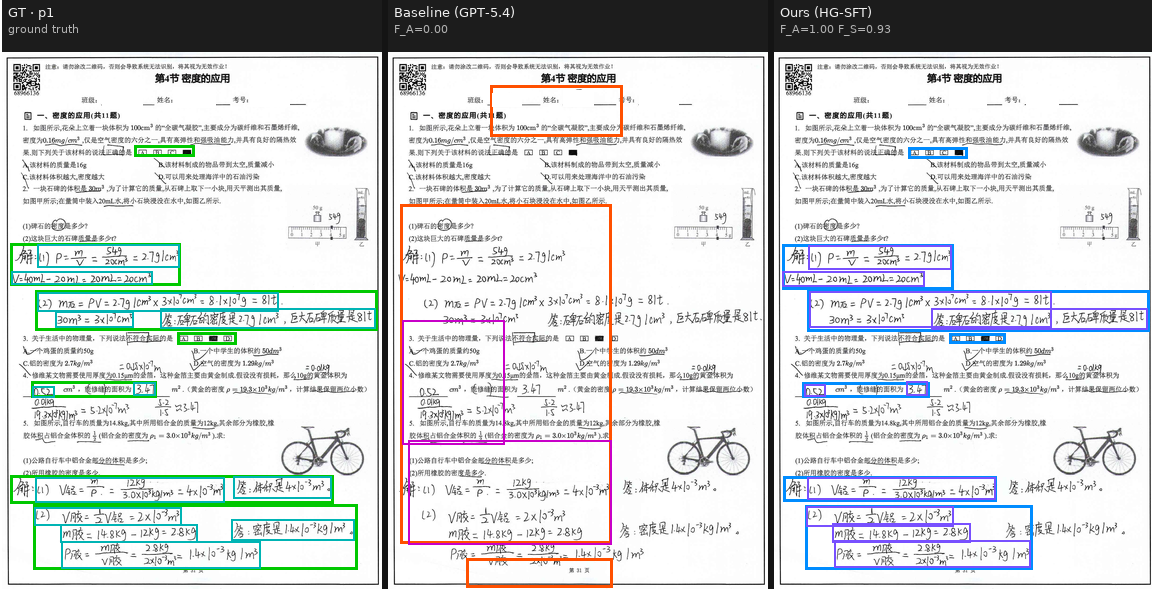}
  \caption{\textbf{Case 1: universally-missed sample (universal rescue).}
  One of the $115$ samples on which no zero-shot baseline reaches
  title-level $\mathrm{F}_1 \geq 0.5$ (Section~\ref{sec:inter-model}).
  Every closed-source frontier API and every open-weight baseline
  produce highly fragmented or mis-localized boxes; the reference SFT
  system recovers both the question-level answer regions and the
  ordered step decompositions.}
  \label{fig:qual-case-hardest}
\end{figure*}

\paragraph{Step-level macro $\mathrm{F}_1$ ($\mathcal{F}_S^{M}$).}
A macro-aggregated complement to $\mathcal{F}_S^{\mu}$. Restricted to the subset of samples that contain at least one step box in ground truth, we compute the per-sample step $\mathrm{F}_1$ (averaging per-page step $\mathrm{F}_1$ within each sample) and then take the unweighted mean over all such samples. Compared with $\mathcal{F}_S^{\mu}$, which up-weights pages with denser step structure, $\mathcal{F}_S^{M}$ treats every step-bearing sample equally and therefore amplifies the contribution of short multi-step answers (single derivations, multi-blank fills). A model that does well only on long derivations but collapses on short multi-step items will show a larger gap between $\mathcal{F}_S^{\mu}$ and $\mathcal{F}_S^{M}$.

\paragraph{Parse success rate ($\mathrm{Succ}\%$).}
The fraction of the $N = 500$ samples on which the model's response (after at most one format-reminder retry) yields a structurally valid JSON array conforming to the prescribed schema. Samples whose response cannot be parsed are counted as failures and contribute $0$ to all localization metrics in the unified score $\bar{\mathcal{S}}$ below.

\paragraph{Unified score over all samples ($\bar{\mathcal{S}}$).}
A reliability-aware composite that combines $\mathcal{F}_A$ and $\mathcal{F}_S^{\mu}$ over the \emph{entire} 500-sample test set: failed-parse samples contribute $0$ rather than being excluded. Concretely, $\bar{\mathcal{S}}$ is the average of the per-sample composite (a weighted combination of question-level and step-level page F1, identical to the per-page reward used during evaluation) over all 500 samples. Comparing $\bar{\mathcal{S}}$ against $\mathcal{F}_A$ exposes the practical cost of format failures: a model with high $\mathcal{F}_A$ but low $\mathrm{Succ}\%$ will see a sharp drop in $\bar{\mathcal{S}}$ (most clearly visible in Doubao-Seed-2.0-Pro at the 2026-02-15 snapshot, where $\mathcal{F}_A = 52.65$ but $\bar{\mathcal{S}} = 21.22$ because nearly half of outputs failed to parse).

\paragraph{Repetition rate ($\mathrm{Rep}\%$).}
The fraction of the 500 outputs in which the model produced a repeating textual or structural pattern (consecutive duplicate boxes, looping JSON fragments, or repeated content blocks detected by a longest-common-substring heuristic on the raw model response). Lower is better. $\mathrm{Rep}\%$ is reported in Table~\ref{tab:main} as a transparency signal; samples flagged as repetitive are still scored under the standard protocol and are not excluded from $\mathcal{F}_A$, $\mathcal{F}_S^{\mu}$, $\mathcal{F}_S^{M}$, $\mathrm{Succ}\%$, or $\bar{\mathcal{S}}$.

\begin{figure*}[ht]
  \centering
  \includegraphics[width=0.95\linewidth]{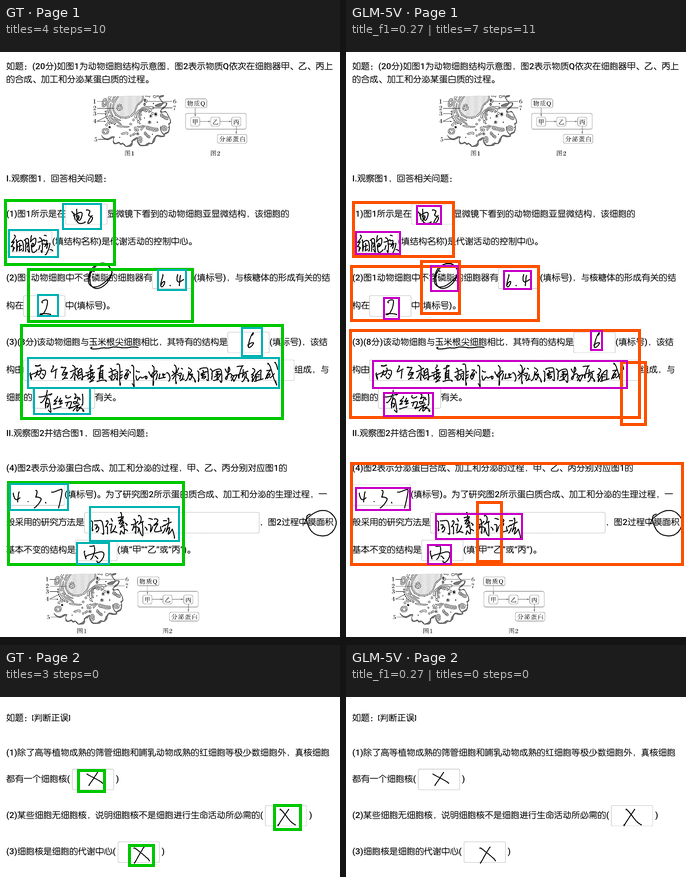}
  \caption{\textbf{Case 2: multi-page enterprise sample.}
  An enterprise answer sheet spanning two pages, illustrating
  page-misalignment and index-shuffling failures: several baselines
  attribute boxes to the wrong page index or emit step IDs out of writing
  order across the page break (cf.\ Section~\ref{sec:bench-breakdown}).
  The reference SFT system preserves page-correct attribution and the
  per-step ordering across pages.}
  \label{fig:qual-case-multipage}
\end{figure*}

\section{Reference SFT System: Full Training Details}
\label{app:sft-hp}

\begin{table}[h]
\centering
\small
\resizebox{\linewidth}{!}{%
\begin{tabular}{ll}
\toprule
\textbf{Item} & \textbf{Value} \\
\midrule
Base checkpoint & GLM-4.6V 9B~\cite{zai2025glmv} \\
Precision & bfloat16 \\
Sequence length & $32{,}768$ \\
Visual tokens / image & up to $10{,}000$ (variable shape, jitter $0.75$--$1.25$) \\
Optimizer & AdamW, $(\beta_1, \beta_2)\!=\!(0.9, 0.95)$, $\epsilon\!=\!10^{-8}$ \\
Weight decay & $0.1$ \\
Gradient clipping & $1.0$ \\
LR schedule & cosine, $30$-step warmup \\
Peak / min LR & $1\!\times\!10^{-6}$ / $5\!\times\!10^{-7}$ \\
Global batch size & $32$ (micro-batch $1$) \\
Training steps & $930$ ($= 3$ epochs over $9{,}920$ examples) \\
Dropout & $0$ \\
\midrule
Hardware & $1$ node, $8\!\times\!\text{H100-80GB}$ \\
Tensor parallel & $2$ \\
Context parallel & $2$ \\
Sequence parallel & enabled (incl.\ ViT tower) \\
Pipeline parallel & $1$ \\
ZeRO / distributed optimizer & enabled, overlapped gradient reduction \\
Activation recomputation & full, per block \\
Activation offload & enabled (incl.\ ViT conv + projection) \\
Optimizer-state offload & CPU, $100\%$, precision-aware moments \\
Wall-clock time & $\approx 3$ hours \\
\midrule
Total compute & $\approx 24$ GPU-hours \\
\bottomrule
\end{tabular}%
}
\caption{Hyperparameters and system configuration of the reference SFT system.}
\label{tab:sft-hp}
\end{table}

\section{Additional Qualitative Cases}
\label{sec:appendix_cases}

This appendix complements the qualitative comparison in Section~\ref{sec:analysis} (Figure~\ref{fig:qual}) with four additional HG-Bench samples that exercise different failure modes and difficulty axes. 

\begin{figure*}[ht]
  \centering
  \includegraphics[width=\linewidth]{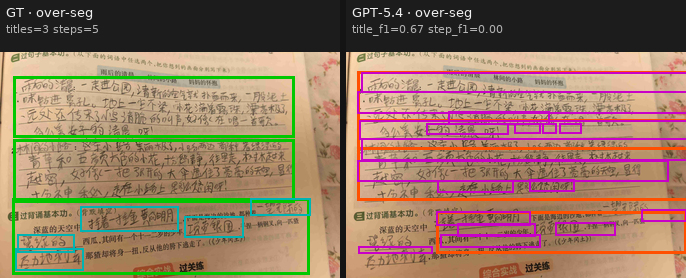}
  \caption{\textbf{Case 3: over-segmentation of step boxes.}
  A single multi-step derivation is split into too many step boxes by
  several baselines (e.g.,\ each ``$=$'' or arithmetic operator gets its
  own box), inflating the step count and confusing downstream per-step
  grading. The reference SFT system produces step boxes whose count and
  granularity match the human annotation.}
  \label{fig:qual-case-over}
\end{figure*}

\begin{figure*}[ht]
  \centering
  \includegraphics[width=\linewidth]{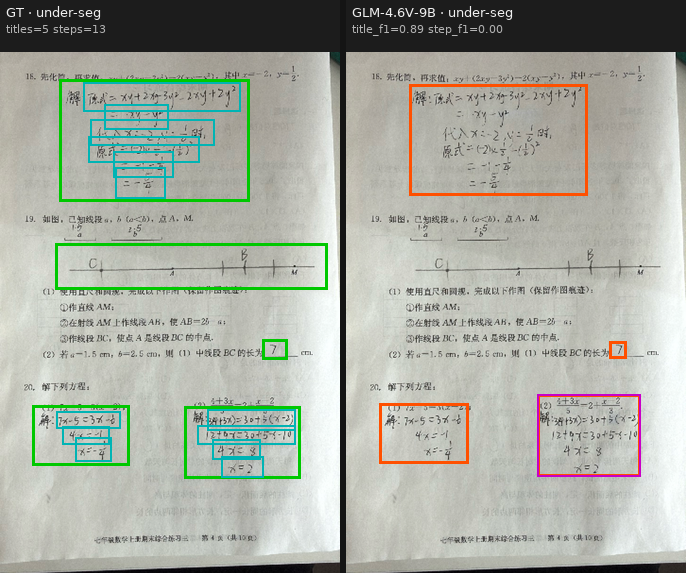}
  \caption{\textbf{Case 4: under-segmentation of step boxes.}
  The complementary failure to Case 3: several distinct derivation steps
  are merged into one large box, losing the per-step grading signal even
  when the overall question-level box is roughly correct. Closed-source
  baselines exhibit this pattern most often on dense multi-line solutions.
  The reference SFT system preserves the per-step boundaries.}
  \label{fig:qual-case-under}
\end{figure*}


\end{document}